\documentclass[10pt,letterpaper]{article}
\usepackage[top=0.85in,left=2.75in,footskip=0.75in]{geometry}

\usepackage{amsmath,amssymb}

\usepackage{changepage}

\usepackage{textcomp,marvosym}

\usepackage{cite}

\usepackage{nameref,hyperref}

\usepackage{amssymb}
\usepackage{stackengine}
\usepackage{lipsum}
\usepackage{multicol}
\usepackage{lineno}
\usepackage{gensymb}
\usepackage{siunitx}
\usepackage{verbatimbox}
\usepackage{longtable}
\usepackage{comment}
\usepackage{graphicx}

\usepackage{lineno}

\usepackage[nopatch=eqnum]{microtype}
\DisableLigatures[f]{encoding = *, family = * }

\usepackage[table]{xcolor}

\usepackage{array}

\newcolumntype{+}{!{\vrule width 2pt}}

\newlength\savedwidth

\raggedright
\usepackage[aboveskip=1pt,labelfont=bf,labelsep=period,justification=raggedright,singlelinecheck=off]{caption}

\makeatletter
\renewcommand{\@biblabel}[1]{\quad#1.}
\makeatother

\usepackage{lastpage,fancyhdr,graphicx}
\usepackage{epstopdf}
\fancyheadoffset[L]{2.25in}
\fancyfootoffset[L]{2.25in}
\begin{document}
\vspace*{0.2in}

\begin{flushleft}
{\Large
\textbf\newline{An Organic Weed Control Prototype using Directed Energy and Deep Learning} 
}
\newline


Deng Cao\textsuperscript{1},
Hongbo Zhang\textsuperscript{2},
Rajveer Dhillon\textsuperscript{3}

\bigskip
\textbf{1} Department of Computer Science, Central State University, Wilberforce, OH, USA \\
\textbf{2} Department of Engineering Technology, Middle Tennessee State University, Murfreesboro, TN, USA \\
\textbf{3} Department of Agriculture Research and Development Program, Central State University, Wilberforce, OH, USA \\

\bigskip



* dcao@centralstate.edu

\end{flushleft}
\section*{Abstract}
 Organic weed control is a vital to improve crop yield with a sustainable approach. In this work, a directed energy weed control robot prototype specifically designed for organic farms is proposed. The robot uses a novel distributed array robot (DAR) unit for weed treatment. Soybean and corn databases are built to train deep learning neural nets to perform weed recognition. The initial deep learning neural nets show a high performance in classifying crops. The robot uses a patented directed energy plant eradication recipe that is completely organic and UV-C free, with no chemical damage or physical disturbance to the soil. The deep learning can classify 8 common weed species in a soybean field under natural environment with up to 98\% accuracy.

\section*{Author summary}
Dr. Deng Cao received his Ph.D in Computer Science from West Virginia University. Dr. Cao joined Central State University in 2013 and currently serves as a Professor of Computer Science in the Department of Mathematics and Computer Science. His research interests include Computer Vision, Machine Learning, Artificial Intelligence, Pattern Recognition, and Advanced Biometrics.

Dr. Hongbo Zhang received his Ph.D in Industrial Engineering from Virginia Tech. Dr. Zhang joined Middle Tennesse State University in 2021 and currently serves as an assisant professor of Mechatronics Engineering in the Department of Engineering Technology. His research interests include Computer Vision, Smart Imaging and Sensing, and Robotics.

Dr. Rajveer Dhillon received his Ph.D in Agricultural Engineering from Virginia Tech. Dr. Dhillon joined Central State University in 2021 and currently serves as a Research Assistant Professor of Precision Agriculture / Agricultural Engineering in the Department of Agriculture Research and Development Program. His research interests include Precision agriculture, Agricultural Robotics, Agriculture Automation, Artificial Intelligence, Internet of Things, and Sensors.


\section{Introduction}

It is estimated that food production must increase 70\% by 2050 to feed 9.6 billion population worldwide \cite{chauhan2020grand}. Crop yield loss due to poor weed control is as high as 32\%, globally. The loss due to weed is higher than the loss from insect pest (18\%) and pathogen (15\%). The application of herbicides for weed control costs about US\$ 40 billion. The costs associated with herbicide resistant weeds also increases. For example, the costs for corn, cotton, and soybeans alone have reached \$1 billion per year. As such, there is a strong demand for weed control.

Numerous organic weed control method exist without the need of herbicide. Hand weeding known effective but impractical for large acreage crop weed control. Biological weeding using biological agents and organisms for weed control \cite{ghosheh2005constraints}. Biological weed control is limited to specific crops, and usually needs to be used with other agents such as herbicides \cite{templeton1990biological}.  Biological weed control is also subjected to strong technology constraints such as shelf life and complex production procedure \cite{boyetchko1997principles}. Flooding is another method of weed control that requires the area being saturated with water a depth of 15 to 30 cm for a period of 3 to 8 weeks for the control of weeds \cite{mcwhorter_1972}. The saturation of the soil reduces the availability of oxygen to the plant roots thereby killing the weed \cite{Rao:2000}. But flooding may adversely impact the yield \cite{uno2021rice}. 

Flame weeding controls weeds with intense heat produced by a fuel-burning device\cite{US5189832}. Yet the poor control of the flaming region makes the method difficult to be used as a robotics weed control solution. Microwave weed control is considered as a pre-sowing weeding method\cite{microwaveweeding}. Similar to flaming weed control, the precision for weed control is problematic when used for robotics weed control \cite{sartorato2006observations}. A number of optical weeding methods were also introduced using laser ablation \cite{Cao:2024} or ultra-violet (UV) light radiation \cite{US8872136}. Similar to flame weeding, laser weeding can be effective but needs to be precisely controlled and directed. Thus, extra human supervising or other safety precautions may be needed in practice \cite{mathiassen2006effect}.

Study shows that UV radiation can cause various types of damage to plants and plant leaves have high absorption coefficient of UV lights \cite{Cline:1966,Barnes:1990}. UV light can be classified into three types: UV-A (315 - 400 nm), UV-B (280 - 315 nm), and UV-C( 100 - 280 nm). Although plant High dose of  UV-B or UV-C exposure may cause damage to plant or even human DNA. However it should be noted that the shorter wavelength UV radiation (UV-B and UV-C) in high dose can  potential cause plant or even human DNA and RNA damages, with significantly increase likelihood of cancer, while longer wavelength (UV-A) radiation is less carcinogenic for cellular structure \cite{Anna:2007}. While effective, these methods are known risky for human health and safety.

In contrast to the risky weed control methods, indigo region illumination (300nm - 500nm) and medium wavelength Infrared radiation (2-20 microns)is applied in this work. The new method is called Rapid Unnatural Dual Component Selective Illumination Protocol (RUDCIP) \cite{US8872136}. One of the dual bands involves UV-A radiation. The UV-A uses a low level of non-mutating UV-A optical energy alongside with near-IR optical energy to create both UV-A below-ground penetration and near-IR impact on foliage above ground. The combination of both approaches yields a very high lethality to the plants by altering cellular metabolism, causing plant damage, hormonal changes, damage to photosynthetic apparatus, and possible interruption of the healthy symbiosis of a plant root with rhizosphere microorganisms surrounding the root. The proposed method does not use high radiative energy transfers for destruction by severe scalding, heat shock or incineration, and is a solution that is safe to humans and animals.

Traditional weed control systems treat the whole field uniformly with the same dose of herbicide. A precision weed control system is selective and therefore needs less herbcide. Selective weed treatment requires the classification of weed versus plant. However, the classification of weed versus plant is not an easy task.  The major challenge is due to the natural environment where occlusion is prevalent. Occlusion of either crop or weed is typical between them or by other natural objects. The complex scene including stone, tree branches, various kinds of grasses is another factor making classification difficult \cite{olsen2019deepweeds}. Conventional methods were designed to classify weeds only rather than classify weeds from other objects such as crops  \cite{olsen2019deepweeds,yu2019deep}. The study of weed and crop classification is  important. Currently, the databases are mostly built for carrots, cauliflower, sugar beet, lettuce, and radish \cite{hu2021deep}. While the databases for large acreage crops such as corn and soybean, those databases tend to be small. For example, 1200 corn  and 400 soybean images \cite{hu2021deep}. Such small-scale databases are not able to offer the robustness of the generalization of crop and weed classification, as such it is not able to be used for robotics weed control.  

With this research, an organic robotics  weed control prototype using directed energy is proposed. A novel weed control-directed optical energy platform was built for organic weed control effective in weed control. Corn and Soybean weed classification databases were built. Deep learning classification methods were used to classify the weed. The results showed the promising performance in classification reaches high accuracy without prior knowledge or preprocessing.

\section{Related Work}

Strong related to the proposed RUDCIP optical directed energy method, laser has been used for weed control. It is known that laser-based weed control is successful. Study showed that the use of laser treatment of apical meristems induced significant growth reduction and lethal effects on weed species \cite{mathiassen2006effect}. More specifically, the laser dot size and exposure time are two deterministic factors for the laser treatment effectiveness \cite{mathiassen2006effect}. Furthermore, study showed that the treament of weed plant with CO2 laser is effective resulting in 90\% of weed fresh mass reduction. The early treatment is crucial where the laser position is also  important \cite{marx2012design}. In contrast to CO2 laser, fiber laser and diode laser however show less effectiveness in controlling weed. Similarly, the treatment location is also important showing the treatment of the meristem is effective \cite{kaierle2013find}. It is further known that the application of middle infrared range laser is effective. The lower power but longer processing time leads to more effective treatment \cite{marx2012investigations}. While laser is effective for weed control, overall, laser is a rather expensive weed control solution and laser safety is a practical safety concern for farmers to adopt the technology.     

To achieve crop and weed classification, Meyer et al. investigated the use of low level texture features such as gray level co-occurrence matrix, angular second moment, inertia, entropy, local homogeneity in soybean, maize and corn fields \cite{Meyer:1999}. Tang et al. propose a texture based weed classification method using the combination of Gabor wavelets and and an Artificial Neural Network (ANN)\cite{Tang:2003}. Romeo et al. propose a fuzzy clustering method for greenness identification \cite{Romeo:2013}. Lottes et al. propose a classification system based on RGB+NIR images for sugar beets and weeds that relies on NDVI-based presegmentation of the vegetation. The approach combines appearance and geometric properties using a random forest classifier \cite{Lottes:2016}. Sujaritha et al. propose a leaf-texture based fuzzy weed classifier in sugarcane fields \cite{Sujaritha:2017}. Recently, CNNs has been used in weed detection and classification due to its high accuracy and robustness. Dyrmann et al. \cite{Dyrmann:2017} present a GoogLeNet based weed detection system. Lottes et al. design a customized CNN for sugar beets and weeds detection that is based on SegNet and Enet \cite{Milioto:2017}. The above studies suggest that with pre-trained networks or customized networks, current weed detection method can achieve high accuracy when the plants are sparsely distributed, but can hardly do so in a field with plentiful plants, mostly because it is difficult to precisely locate a plant when it is very close to or tangled with other plants.

 Deep learning has been used to detect plant diseases and weed identification and has shown great success in many specific cases \cite{Shin:2016, Andreas:2018, Zhang:2023100123,Mouradantas:2023}. As such, based on the soybean and corn databases, a transfer learning approach is used for the proposed research. In this work, a number of pre-trained networks, including AlexNet\cite{alexnet}, Squeezenet\cite{squeezenet},GoogLeNet\cite{googlenet}, Inception v3\cite{inceptionv3}, Densenet\cite{densenet}, MobileNet\cite{mobilenetv1, mobilenetv2}, resNet\cite{resnet}, Xception\cite{xception}, Inceptionresnet\cite{inceptionresnetv2}, shufflenet\cite{shufflenet} and Nasnetmobile\cite{nasnet} are retrained on our database using transfer learning. An end-to-end weed classification system is then built based on the re-trained networks. The system is implemented using MATLAB 2019a and TensorFlow 1.7. The classification performance are compared and discussed.

\section{Approach/Methodology}\label{dissb}

\subsection{Prototype Phase 1}
The phase one of the prototype design of the directed energy used a towe structure. The Distributed Array (DA) unit housing the RUDCIP optical directed energy can be towed behind a tractor or an all terrain robot that has an adjustable height hitch. The DA has 7 x 15 36V/410W light bulbs as the source of directed energy. Each bulb has a power output of 5.73W per $cm^2$ (reading taken by ThorLabsPM10OD power and energy meter with S314CC sensor). The UV-A output for each bulb is 11mW per $cm^2$. The light bulbs are mounted in 4 x 4 inches (102 x 102 mm) reflectors. The reflectors are 6 inches above the ground.

The RUDCIP Band specification is generated by a preponderance of data obtained empirically by experiments upon a specific plant species known difficulty in eradication and control, notably being Red River Crabgrass (Digitaria cilaris). The RUDCIP Band specification is formed based upon a specific protocol specification. It is produced based on weed control experiment of Red River Crabgrass. The details of the experiment is specified in ~\ref{eqn:RUDCIP}. Based on the experiment, the following lethality effectiveness equation is obtained showing a linear relationship between lethality and treatment duration consistent to previous literature \cite{mathiassen2006effect} . 

\begin{equation}
\label{eqn:RUDCIP}
L=5.5 \times 10^{-6} \times E_{near\mbox{-}IR} \times T_{near\mbox{-}IR} + 6.5 \times 10^{-5} \times E_{UV\mbox{-}A} \times T_{UV\mbox{-}A},
\end{equation}

where L is unitless nominal lethality effectiveness expressed in fraction of plants dead in 30 days, such that L is greater than zero and equal to or less than one. $E_{near\mbox{-}IR}$ and $E_{UV\mbox{-}A}$ are radiation of irradiance in $W/m^2$, $T_{near\mbox{-}IR}$ and $T_{UV\mbox{-}A}$ are total exposure time in seconds.

\begin{figure}[htp]
\begin{center}
\includegraphics[height=1.4in]{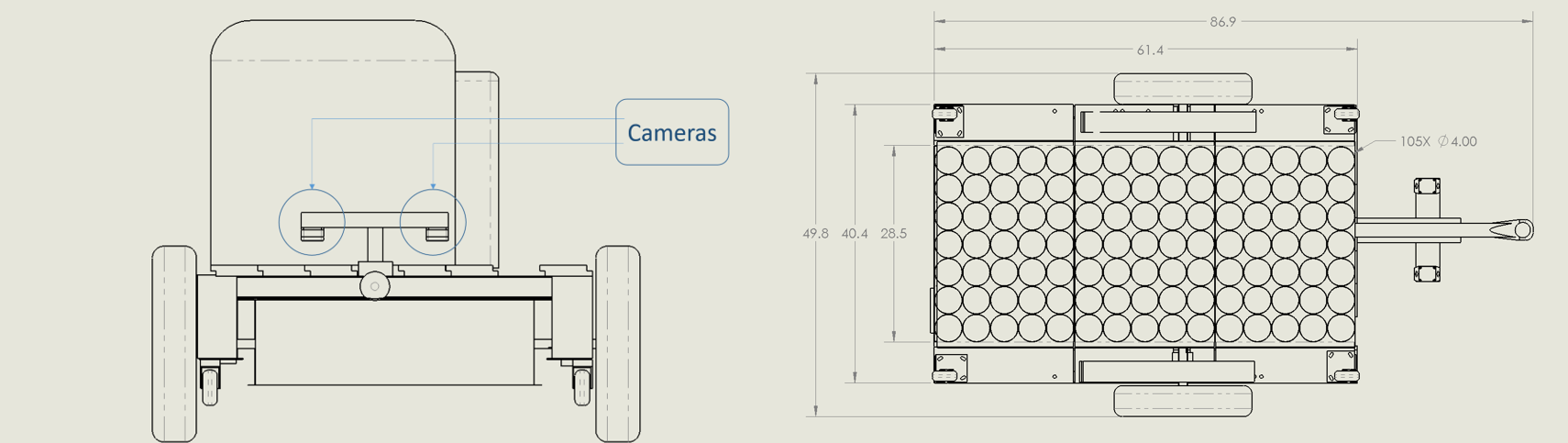}
\end{center}
\caption{Left: The front view of the DA. Right: The bottom view of the DA.}
\label{fig:DA2}
\end{figure}

Two COTS Logitech HD webcam C270 are mounted at the front bar of the DA, 20 inches above the ground. The cameras feature video capture up to 1280 x 720 pixels and still image capture up to 3.0MP. In experiment, the images are taken under video model using a 640 x 480 resolution. For the convenience of manual labeling, the images in the database are resized to 1280 x 960. The cameras are carefully mounted and aligned so that a pair of images from the left camera and right camera can cover the length of a full row of 7 reflectors (see Fig ~\ref{fig:DA2}).


\begin{figure}[htp]
\begin{center}
\includegraphics[width=4in]{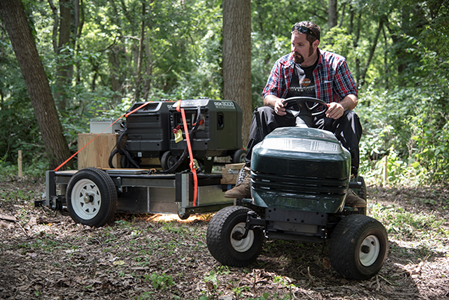}
\end{center}
\caption{The Demonstration of DA at WPAFB test site.}
\label{fig:DAwithTractor}
\end{figure}

\begin{figure}[htp]
\begin{center}
\includegraphics[width=4in]{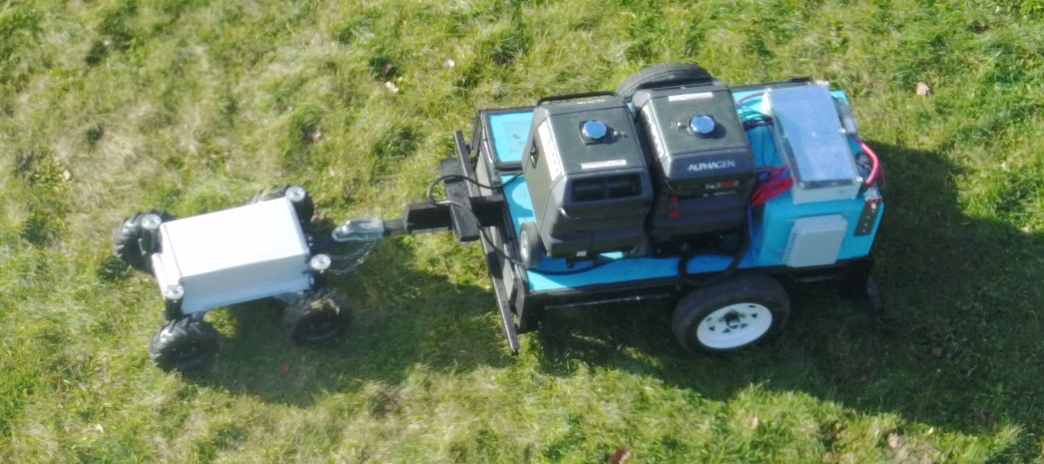}
\end{center}
\caption{The DA towed by a WC800-DM4, 4WD All Terrain Robot.}
\label{fig:DAwithDroid}
\end{figure}

The DA is developed using a “Move then Dwell” approach for weed control. The image recognition is used to recognize the weed. Upon identification of the weed, the robot will be paused. Subsequently, the directed energy sources are turned on for weed control. Currently, the device can support up to 6400W, so a maximum of 16 directed energy sources can be turned on at any one time. Two sets of 16 lights are used for the treament areas until  the area is covered by 105 directed energy lights. The DA can be towed by a regular tractor and the weed control performance has been verified at Wright-Patterson Air Force Base (Fig ~\ref{fig:DAwithTractor}). The DA can also be towed by a 4WD all terrain robot (Fig ~\ref{fig:DAwithDroid}). The robot is built based on a WC800-DM4 Robot platform with the size of 36.5 x 28.9 x 18.4 inches (927 x 734 x 467 mm) that can be found on superdroidrobots.com. The robot also has two Logitech HD webcam C270 mounted at the front bar, with the same setup as the ones mounted on the DA. 


The phase I prototype has the following limitations. The directed energy recipe takes up to 30 seconds irradiation time to achieve high empirical lethality, which is time consuming comparing to other organic weeding methods. The power consumption is relatively high and only 15 reflectors can be used. This prototype does not consider the in-row and between-row differences in the actual crop field. Since it has two separated parts (the towing robot and the DA unit) and the towing robot can wiggle when driving, a mismatch could occur between calculated weed locations and actual weed locations. Although this issue can be rectified by mounting sonar sensor, accelerometer and gyroscope to track deviations and perform course correction, a simpler solution would be mounting the cameras and reflectors closer to each other, which requires redesigning the structure of the DA and the position of the cameras.

\subsection{The Phase II prototype}

\begin{figure}[htp]
\begin{center}
\includegraphics[width=4in]{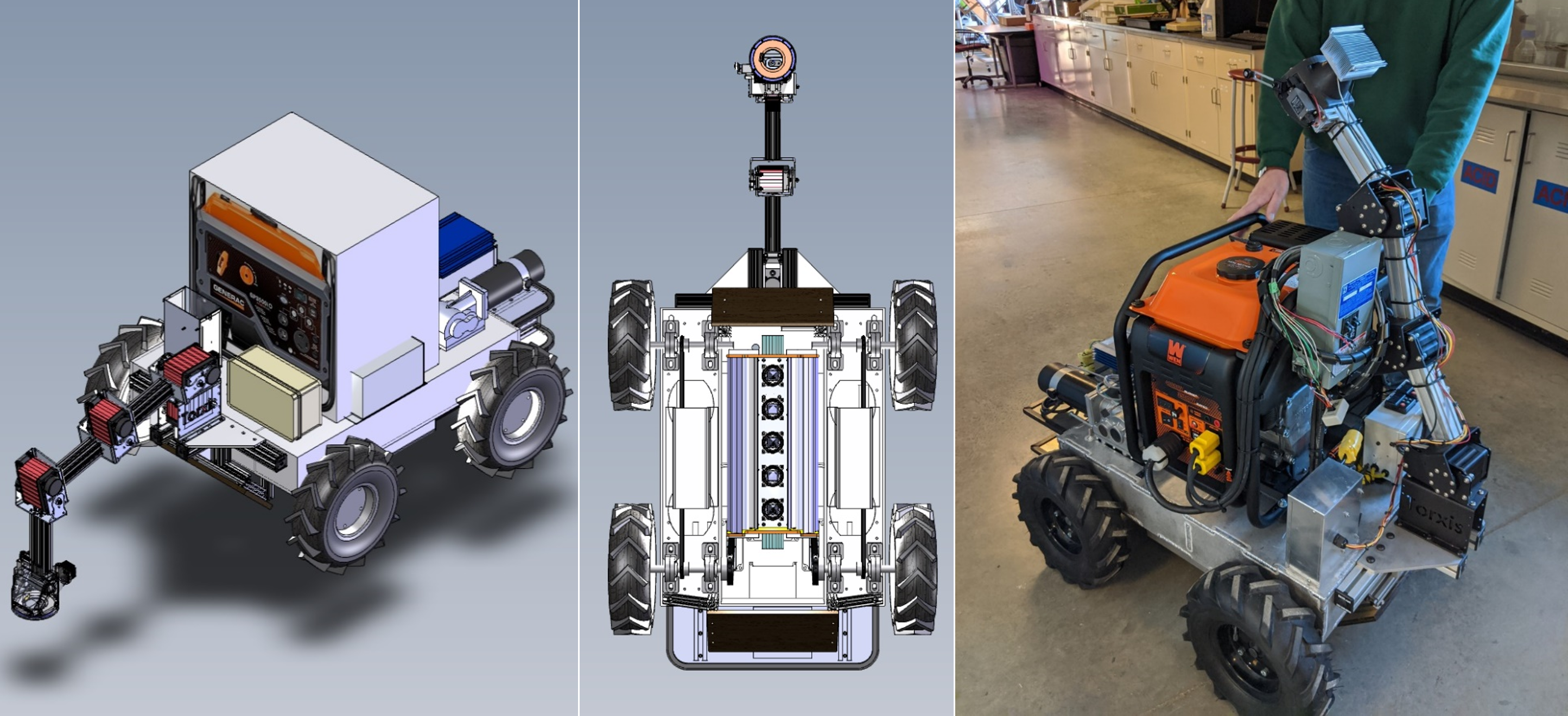}
\end{center}
\caption{The proposed prototype Disruptor Array Robot version 4 (DAR4)}
\label{fig:DAR4}
\end{figure}

The phase II prototype is designed for small farms. The proposed prototype, Disruptor Array Robot version 4 (DAR4) is built from a WC400-DB4 4WD All Terrain Robot Platform from superdroidrobots.com. It is equipped with one LED directed energy reflectors mounted on its articulating Torxis modular arm and five to its bottom. It also equipped with a 1280 x 720 camera with real time video streaming and LED illumination. Its all-in-one architecture provides better synchronous image processing and reduces the mismatching issue we had in the phase I model (Fig ~\ref{fig:DAR4}).


\begin{figure}[htp]
\begin{center}
\includegraphics[width=4in]{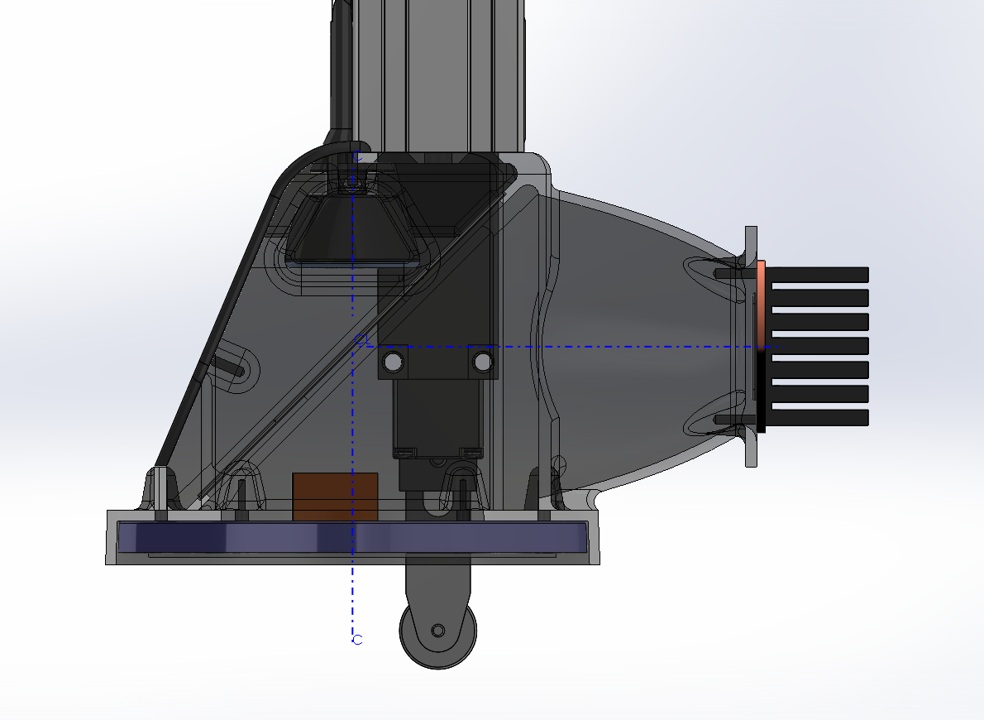}
\end{center}
\caption{The end effector of the robotic arm of DAR4.}
\label{fig:endEffector}
\end{figure}

The end effector (Fig ~\ref{fig:endEffector}) of the robotic arm has a novel design with a built-in 1280 x 720 camera and a 100W kapton heater mounted to a plastic ring. A  borosilicate glass (pyrex) is attached to the cover to protect the camera. When operating, the pyrex is pre-heated to and maintained at 400F. A 100W, 450nm LED array from Chanzon  emitting light through the heated pyrex and to the target weed for weed control. An optical filter is located at 45 degrees to the ground behind the pyrex. It is used to reflect the LED light towards the weed meanwhile allowing the camera to visualize the scene. The size of optical aperture is adjusted and positioned so that it does not interfere with the operation of the blue LED’s and the camera. 

\begin{itemize}
  \item {$0.06W/cm^{2}$ of mid-wavelength infrared (MWIR)}
  \item {$0.85W/cm^{2}$ of 450nm indigo region illumination distributionblue light (IRID)}
\end{itemize}


\subsection{Development of Soybean and Corn Databases}
\subsubsection{The Soybean Database}

For classification of soybean and weed, soybean weed database was created. In the database, it consists of 3,000 375 x 375 images. The images are cropped from larger raw images collected from a soybean field outside the university between 2017 and 2021. 
The size of the image is based on the the prototype of the underbody 1 x 5 grid and each cell in the grid corresponds to a directed energy source. Each DE source is a 4 x 4 inches (102 x 102 mm) reflectors with. The database will help to understand  that  if there is a weed in each cell. If there is a weed, the cell is activated. An AC power ACR source will be switched on for weed control.

The database images are then manually categorized into two classes: Weed and Non-Weed. The non-Weed class contains 1,500 images of soybean seedling in Cotyledon (VC), First trifoliate (V1), Second trifoliate (v2), third trifoliate (V3), fourth trifoliate (V3), and fifth trifoliate (V5) stages, along with background (soil, rubble or dead plant) images. The Weed class contains 1,500 images of 8 common Ohio weed species: Dandelion	(Taraxacum officinale F.H.Wigg.), Hairy Crabgrass (Digitaria sanguinalis L.), Eastern Cottonwood (Populus deltoides W. Bartram ex Marshall), American Pokeweed (Phytolacca americana L.), Broadleaf Plantain (Plantago major L.), Buckhorn Plantain (Plantago lanceolata L.), Carpetweed (Mollugo verticillata L.) and Yellow Woodsorrel (Oxalis stricta L.).

The database is also diversified including a variety of images including sunny or cloudy weather condition, different land conditions of dry, wet, or muddy. Plants in the database are either clean or covered with dirt. Some soybeans are with broken leaves, and some are dead (after treatment).  Shadows are also included in the database. Images with both objects from both classes (crop and weed) are not selected.

\subsubsection{The Sweet Corn Database}
The Sweet Corn database was established between 2019-2021. It consists of 2,500 RGB images of various sizes in three classes. The Sweet Corn (Zea mays L. var. rugosa Bonaf.) class containing 628 images of sweet corn in their vegetative stages from emergence (VE) to V4. The Weed class contains 1,177 images of the common US weed species as described in the soybean database. And the Soil class contains 695 images of common soil and dirt. Fig ~ref{fig:databaseSample} shows samples of these plant species.


\begin{figure}[htp]
\begin{center}
\includegraphics[width=5in]{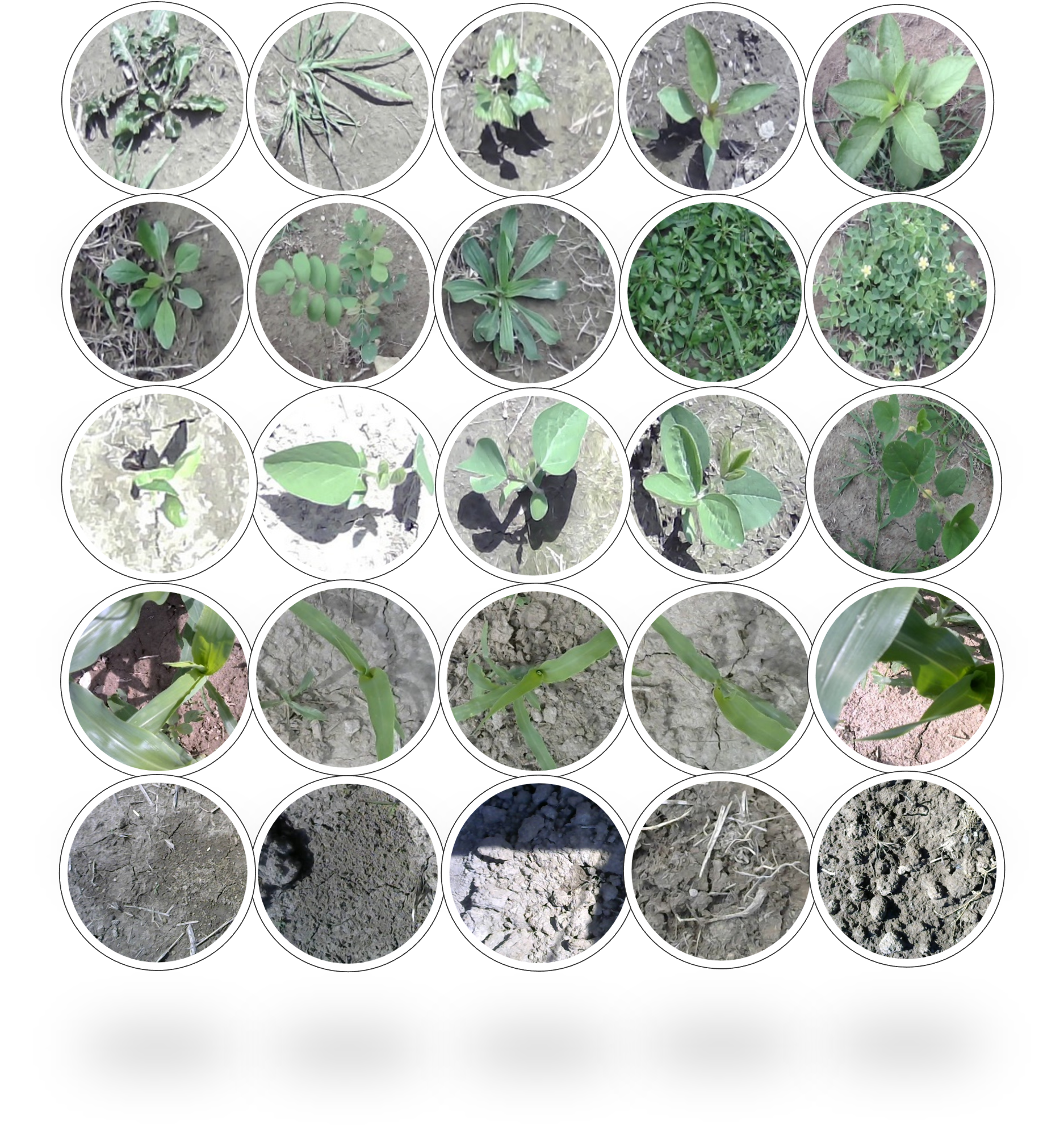}
\end{center}
\caption{1st and 2nd line: common Ohio weed species; 3rd line: soybean seedlings; 4th line: sweet corn seedlings; 5th line: common soil.}
\label{fig:databaseSample}
\end{figure}

\subsection{Network architecture}
Following the creation of databases, we experimented the use of convolutional neural networks for classification.  Through the training, in order to make a network exceptional robust to changing environments factors, the deep network needs to be trained with a large amount of diversified data from a wide variety of illumination and weather conditions, growth stages, and soil types, if trained from scratch. Unfortunately, this comes at a high cost and we therefore propose a transfer learning that utilizes the pre-trained networks given the limited training data.

On the other hand, transfer learning is commonly used in deep learning applications. Fine-tuning a network with transfer learning is usually much faster and easier than training a network with randomly initialized weights from scratch. The pre-trained features can be quickly transferred to a new task using a much smaller number of training images. For instance, GoogLeNet is a 22-layer deep network well known for its relatively speed and robustness. In order to retrain GoogLeNet for weed classification, the last three layers with the names 'loss3-classifier', 'prob', and 'output', are removed from the GoogLeNet and three new layers, a fully connected layer, a softmax layer, and a classification output layer are added to the layer graph. The final fully connected layer is set to have the same size as the number of classes in the new databases.

\section{Results}\label{results}
The performance of the weed classification system based on CNN is explained in this section. The accuracy $x$ of the classification system is measured by using the well-known and widely used formula, Eq~\ref{eqn:accuracy},

\begin{equation}
\label{eqn:accuracy}
x = \frac{TP+TN}{TP+TN+FP+FN}
\end{equation}

where $TP$ is the number of True Positives, $TN$ is the number of True Negatives, $FP$ is the number of False Positives and $FN$ is the number of False Negatives.

80\% of the images in the soybean database are randomly selected as the training set, and the rest 20\% of the images are used as the test set. The test data is used only once for reporting the accuracy. Similarly, an 90\% - 10\% split is used for training set and test set respectively for the sweet corn database. For each neural network, the experiment is repeated $n$ times ($n = 10$ in this case) on each network and the average classification accuracy $\bar{x}$ is calculated as in Eq~\ref{eqn:final accuracy}:

\begin{equation}
\label{eqn:final accuracy}
\bar{x} = \sum_{i=1}^{n} \frac{x_{1}+x_{2}+...+x_{n}}{n}
\end{equation}

\begin{figure}[htp]
\begin{center}
\includegraphics[width=5in]{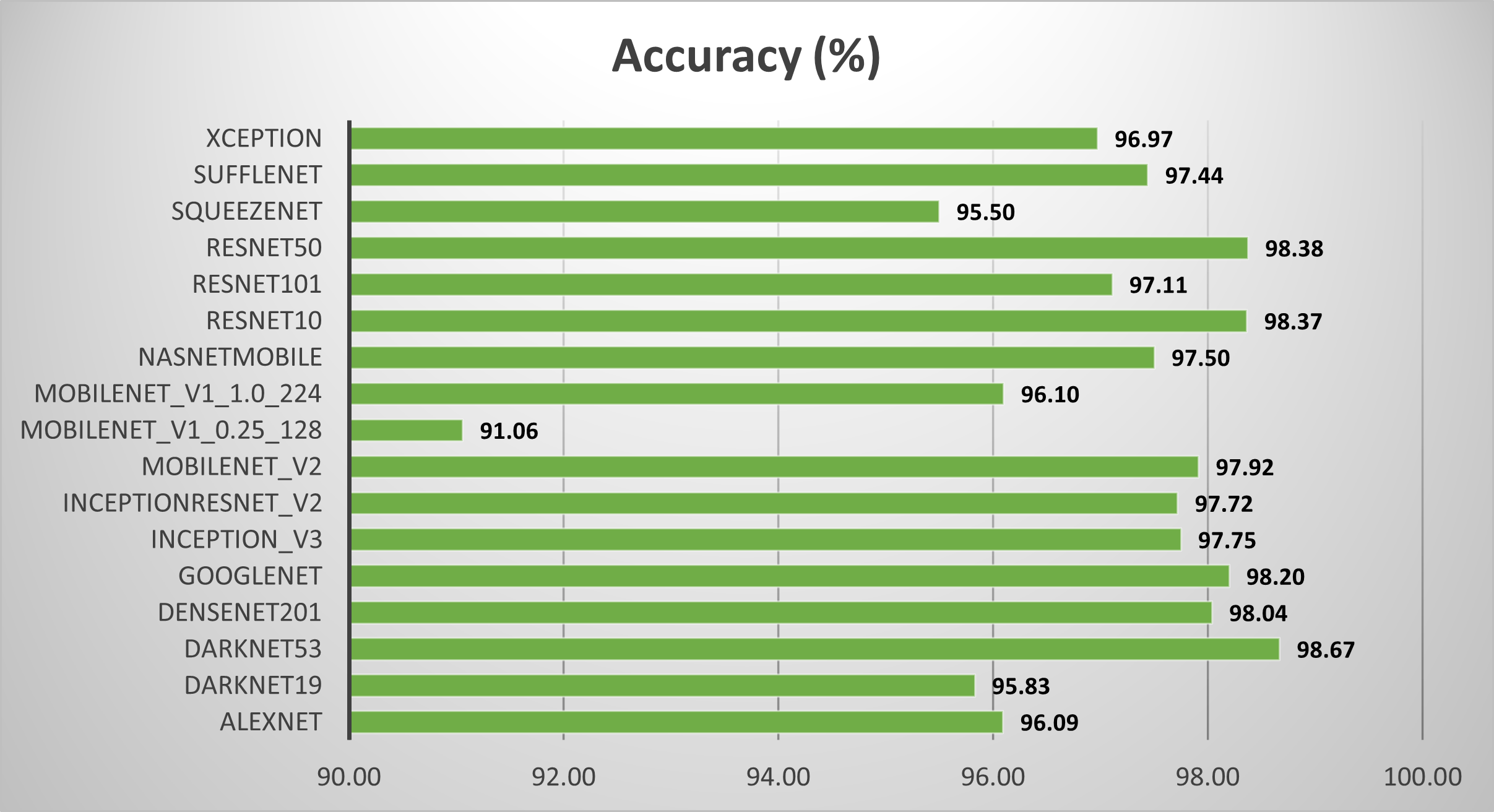}
\end{center}
\caption{Comparison of accuracy and sizes of tested networks.}
\label{fig:resultsSoybean}
\end{figure}

\begin{figure}[htp]
\begin{center}
\includegraphics[width=4in]{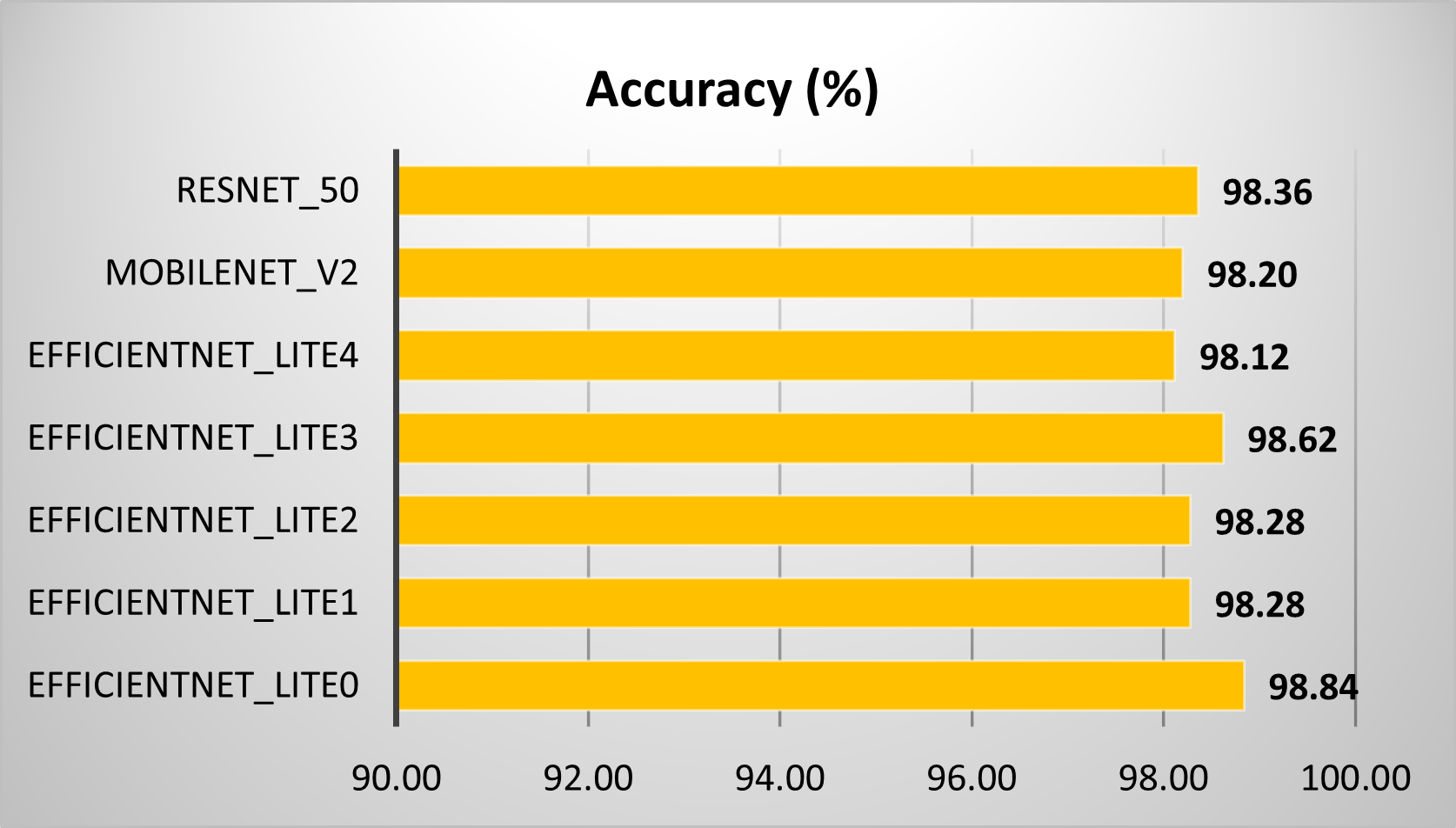}
\end{center}
\caption{Comparison of accuracy and sizes of tested networks.}
\label{fig:resultsCorn}
\end{figure}

The experimental results suggest that the tested pre-trained networks are transferable to crop/weed/soil classification task without loosing the performance (Fig ~\ref{fig:resultsSoybean} and Fig ~\ref{fig:resultsCorn}). Most of the tested networks reached over 95\% accuracy and some reached over 98\% accuracy. Some light weighted networks such as MobileNet\_v2 or Efficient\_lite can easily achieve process speed around 20 - 30 frame per second on a regular tablet or laptop without GPU support, which is sufficient for the real-time weeding application.


\section{Conclusion}
In this work we proposed a non-chemical, non-invasive weed control prototype that is mainly designed for small organic farms. The prototype uses CNN for weed identification and directed energy for organic weed control. The weed identification system was evaluated on two outdoor crop databases for its effectiveness. The results showed classification accuracy up to over 98\%, which is comparable to the current state-of-the-art image classification accuracy in general.

\section*{Discussion and Future Work}

The previous prototype does have certain limitations. The directed energy recipe takes up to 30 second irradiation time to achieve high empirical lethality, which is time consuming comparing to other organic weeding methods. When weeds grow larger than 4 x 4 inches, 100\% lethality is not guaranteed. In order to address this issue, a new LED based recipe is under development and the goal is to reduce the treatment time to 2-5 seconds and reduce the energy consumption by 70\%. Within that time, the weed can be treated by a number of directed energy sources in sequence when the DA is moving. If the DA moves at 1km per hour, it will give 6 seconds for the target weed to be treated by 15 directed energy sources. The new recipe will also be a safer solution because the LED will generate less heat.

Another issue is that the weed classification and weed treatment are not happening simultaneously. In fact, the CNN always check an area that is in front of the DA, not the area that is under the DA. The tractor or robot can wiggle when driving, which can cause mismatching between expected weed locations and actual weed locations. Although this can be rectified by mounting sonar sensor, accelerometer or gyroscope to track deviations and perform course correction, a simpler solution is to mount the cameras and reflectors close to each other, which requires redesigning the structure of the DA and the position of the cameras. Such new DA model is under development.

The third issue of this work, is how to make the towing robot autonomous so that the weeding can be done regularly while the owner is on vacation. For a small and slow-moving vehicle that only works in the crop field, this goal can be achieved by equipping the robot with 2D simultaneous localization and mapping (SLAM) system \cite{Hess:2016} and object detection based accident avoidance system \cite{Sharma:2017}.

\section{Acknowledgments}
This work is supported by the following grants and local business agency:

\begin{itemize}
  \item {USDA Capacity Building Grant: All-in-One Organic Weed and Crop Disease Management Using Directed Energy and Convolutional Neural Networks, 2019-38821-29152, 2019 - 2024.}
  \item {USDA Evans-Allen program: Advanced Agriculture Technologies for Small Scale Farms, 2022 - 2027.}
  \item {Google TensorFlow College Award, 2022.}
  \item {Global Neighbor, Inc.}
\end{itemize}
\nolinenumbers

\bibliography{Thesis_Deng_2023}


\end{document}